\documentclass{article}
\usepackage{ijcai16}

\usepackage{times}
\usepackage{epsfig}
\usepackage{epstopdf}
\usepackage{graphicx}
\usepackage{amsmath}
\usepackage{amssymb}
\usepackage{algorithm}
\usepackage{algpseudocode}
\usepackage[table]{xcolor}
\usepackage{enumitem}
\usepackage{multirow}
\usepackage{flushend}
\usepackage{tabu}
\usepackage{color}
\usepackage{array}
\usepackage{url}
\usepackage{pifont}

\newcommand\Mark[1]{\textsuperscript#1}

\algrenewcommand\algorithmicindent{1.0em}%
\renewcommand{\algorithmicrequire}{\textbf{Input:\:}}

\title{Balancing Appearance and Context in Sketch Interpretation}

\author{
Yale Song\Mark{1}, 
	Randall Davis\Mark{2}, 
	Kaichen Ma\Mark{3}, 
    Dana L. Penney\Mark{4} \\
\Mark{1}Yahoo Research, 
	\Mark{2}Massachusetts Institute of Technology, \\
    \Mark{3}Google Inc., 
    \Mark{4}Lahey Hospital and Medical Center
\thanks{This work was done when Y. Song and K. Ma 
  		were at MIT Computer Science and Artificial 
        Intelligence Laboratory.}
}

\begin{document}

\maketitle

\begin{abstract}
We describe a sketch interpretation system that detects and classifies clock numerals created by subjects taking the Clock Drawing Test, a clinical tool widely used to screen for cognitive impairments (e.g., dementia). We describe how it balances appearance and context, and document its performance on some 2,000 drawings (about 24K clock numerals) produced by a wide spectrum of patients. We calibrate the utility of different forms of context, describing experiments with Conditional Random Fields trained and tested using a variety of features. We identify context that contributes to interpreting otherwise ambiguous or incomprehensible strokes. We describe ST-slices, a novel representation that enables ``unpeeling'' the layers of ink that result when people overwrite, which often produces ink impossible to analyze if only the final drawing is examined. We characterize when ST-slices work, calibrate their impact on performance, and consider their breadth of applicability. 
 \end{abstract}

\section{Introduction}
Our real-world sketch interpretation task comes from the Clock Drawing Test, a deceptively simple test that has been used for more than fifty years to help determine cognitive status. This is a task of growing importance given the ``greying'' of populations around the world and the increasing impact of cognitive impairments~\cite{alzheimer20132013}. The test asks a subject to draw on a blank sheet of paper a clock face showing a particular time, then asks them to copy a pre-drawn clock shown on another sheet. Decades of experience with the test have made it a widely used screen for cognitive impairment of many sorts, e.g., Alzheimer's, dementia, stroke, etc.~\cite{solomon19987}. 

Interpretation of the test is traditionally based on what the patient drew (e.g., are all the numbers present at correct locations) and how accurately it was drawn (e.g., how closely do the hands point to the correct numerals). Interpretation is currently entirely manual, requires extensive clinical experience and knowledge, is at times labor-intensive, and has low inter-rater reliability. These issues limit its clinical effectiveness, particularly when the goal is detecting impairment early in disease development.

\begin{figure}[tp]
	\begin{center}
		\includegraphics[width=.48\textwidth]{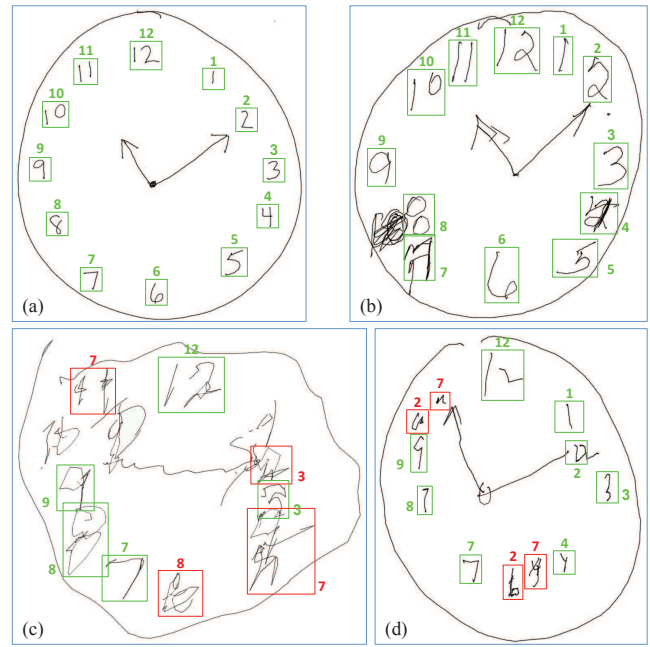}
	\end{center}	
	\caption{Our task is to detect and to classify clock numerals from drawings created by subjects taking the Clock Drawing Test. Shown here are (a) simple example (healthy subject), (b) overwritten, (c,d) impaired subjects, and our numeral detection/classification results (green: correct, red: incorrect).}
	\label{fig:fig1}
\end{figure}

Since 2006, our research group has been conducting the THink project~\cite{DavisLAPP15}, administering the digital Clock Drawing Test (dCDT) using a digitizing ballpoint pen (from Anoto Inc.). The pen records its position on the paper with considerable precision ($\pm$ 0.002 inches) every 13.3 ms, providing a sequence of time-stamped coordinates. The data thus captures both the drawing and the behavior that created it. We have accumulated a database of several thousand tests from both healthy and cognitively impaired subjects, a unique dataset of real-world drawing behavior. We have established and recorded ground truth labels for each pen stroke (i.e., which numeral it is, which hand, etc.), allowing us to calibrate the performance of our stroke classifiers.

Our task in this paper is a slightly unusual form of sketch interpretation: we know from the outset what the user is trying to draw -- a clock. The challenge is to understand what they actually produced, i.e., determine the right label for every stroke. Correctly labeled strokes are crucial to the ultimate aim of this work: automated screening (disease vs healthy) and diagnosis (disease selection). Recent work has shown that correctly labeled strokes enable both of these to be done with demonstrably higher performance than existing algorithms~\cite{souillard2015learning}. We focus here on clock numerals because numeral features play a particularly important role in clinical interpretation of the test.

Our task is clearly not traditional isolated digit recognition (e.g., MNIST~\cite{lecun1998mnist}), as we have to determine which strokes are likely to be the digits, and where they are. It also is considerably more difficult than most digit recognition because impaired users at times draw digits that cannot possibly be interpreted accurately in isolation.

Despite its domain-specificity, the task presents interesting challenges widely applicable to sketch recognition. First, our system must deal with the range of phenomena produced by a population that is in some cases impaired and is in all cases completely naive. Our users are the furthest thing from trained users: A fundamental premise of the test is in fact that it captures a person's normal, spontaneous behavior. (This is one reason why we use the digitizing ballpoint; a tablet could distort the results by its different ergonomics and novelty, particularly for older users.) 

Second, our system must handle corrections in natural handwriting. When users correct mistakes, they often do so by crossing out and over-writing, adding to the difficulty of interpretation, sometimes making recognition impossible to do from the image alone (e.g., the overwritten regions in Fig.~\ref{fig:fig1} (b)). Finally, our system must be prepared for elements that are distorted, misplaced, repeated, or missing entirely (Fig.~\ref{fig:fig1} (c) and (d)), as our users range in age from their 20's to well into their 90's, and span from completely healthy to those with cognitive impairments.

\section{Our Approach}
Our system starts its stroke recognition by finding an initial approximation of the center of the drawing, for use as a reference point for angular measures. This is frequently provided by identifying the clock circle, typically the longest circular stroke(s). We find a best-fit ellipse to the points in these strokes, giving both a good initial approximation to a center and a measure of clock size. If there is no clock circle (as can happen), we use the center of mass of all of data points.

\subsection{Digit / Non-digit Stroke Classification}
The first step is to separate digit stokes from non-digit strokes. We expect digits to be further from the clock center, and have observed a tendency for users to complete the task in categories: clock circle, numerals, hands (not necessarily in that order). There is thus both a temporal and a spatial dimension in which digit strokes may be grouped. Accordingly, we use k-means clustering~\cite{arthur2007k} in a 2-D space defined by stroke starting time and distance from the clock center.

The result is three groups of strokes, one intended to contain the clock circle, another intended to contain hand strokes, and the last intended to contain digits and other strokes sharing the same space/time region, e.g., tick marks. Since the majority of strokes represent numerals in a clock drawing, we consider the largest cluster the digit cluster and focus here on those, discarding others.

\subsection{ST Segmentation of Digit Strokes}
\label{sec:segment}

\begin{figure}[tp]
	\begin{center}
		\includegraphics[width=.48\textwidth]{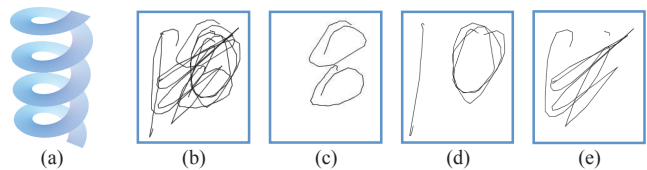}
	\end{center}	
	\caption{An illustration of ST-slices: (a) a helical ramp; (b) the final ink excerpt from Fig.~\ref{fig:fig1}b near the 8 position; (c, d, e) the final ink unpacked into ST slices.}
	\label{fig:st-slice-example}
\end{figure}

In terms of the original drawing, the set of presumed digit strokes lie in a very roughly annular area. The next task is to divide the annulus into segments intended to contain individual clock numerals (1 to 12). In a clean drawing, a simple angular difference threshold works well (e.g., Fig.~\ref{fig:fig1} (a)). But an angular measure would not work for drawings like Fig.~\ref{fig:fig1} (b), (c) or (d); in fact it is impossible to tell from the drawing alone what happened near the 8 position in Fig.~\ref{fig:fig1} (b).

As our data is time-stamped, we can (literally) play a movie of the test, a remarkably effective way of enabling a person to make sense of complex, overwritten ink. 

Wanting the computer to similarly unravel the layers of ink motivated a hybrid representation combining space and time. A spatial representation alone is inadequate because of overwriting, which can produce incomprehensible ink (e.g., Fig.~\ref{fig:fig1} (b)). A temporal representation alone is inadequate because users may go back and add strokes to areas where they drew previously; these additional strokes make sense only in the context of what was drawn there (perhaps much) earlier.

We thus segment the annulus by considering both space and time. For any pair of strokes in time sequence, we compute their central angle (with respect to the estimated center of a clock) and their time difference, producing a 2-D feature vector. We then examine each pair sequentially, and segment them using a boosted logistic regression classifier~\cite{friedman2000additive} trained on a held-out dataset, annotated with binary labels indicating the ground truth segments.

We call the resulting representation spatio-temporal slices (ST-slices) and find that they are an effective representation for untangling the ink in complex drawings. One way to make the idea more intuitive is to imagine the digits being drawn on a helical ramp of paper  (Fig.~\ref{fig:st-slice-example} (a)), where the vertical dimension is time. We segment the ramp whenever two consecutive strokes are sufficiently far apart angularly and/or temporally according to the segmenter. The result is a collection of ST-slices, each holding a set of strokes that is likely to represent a single clock numeral.

The clock in Fig.~\ref{fig:fig1}(b) provides one example of the effectiveness of the ST-slice representation. The ST-slices produced near the 8 position of the clock reveal what is otherwise hidden: As shown in Fig.~\ref{fig:st-slice-example}, there was first an ``8'' drawn in the appropriate position on the clock (Fig.~2(c); undetectable in the final drawing). The 8 was later overwritten by a 10 (Fig.~2(d)); both of these were later scratched out (Fig.~2(e)). (The ``8'' clearly visible in Fig.~\ref{fig:fig1}(b) is a distinct pair of strokes added later next to the over-writes.) 

In terms of the ST-slices, each of these drawing actions appears on a separate layer of the helical ramp, with their overlapping angular position on the clock face captured in their matching angular positions on the ramp. ST-slices are effective in ``unpeeling'' layers of ink produced when people overwrite. The resultant ability to see into the layers and interpret overwritten characters appears to be unique to our work. They are also ``low cost,'' in the sense that they default to ordinary angular slicing in the absence of overwriting.

\subsection{Overwriting and Augmentation}
\label{sec:overaug}
\begin{algorithm}[tp]
\caption{Detecting Overwriting and Augmentation}
\label{alg:overaug}
\begin{algorithmic}[1]
\State \algorithmicrequire{Chronologically ordered set of ST-slices $\mathcal{S}$}
\Procedure{Overwriting\textendash Augmentation}{$\mathcal{S}$}
  \For{$(s_{i},s_{j}) \in \mathcal{S}, j > i$ }
    \If{overlap($s_{i},s_{j}$) $> \theta_{1}$} 
      \State // Deemed overwrite: $s_{j}$ overwrites $s_{i}$
      \State $\mathcal{S} \leftarrow \mathcal{S} \setminus \{ s_{i} \}$
	\ElsIf{overlap($s_{i},s_{j}$) $> \theta_{2}$} 
      \State // Deemed augmentation: $s_{j}$ augments $s_{i}$
      \State $s_{i} \leftarrow \mbox{merge}(s_{i},s_{j}) $
      \State $\mathcal{S} \leftarrow \mathcal{S} \setminus \{ s_{j} \}$
    \EndIf
  \EndFor
\EndProcedure
\end{algorithmic}
\end{algorithm}

When ST-slices overlap spatially, we look at pairs of those slices to determine whether they represent overwriting (the second set of strokes is intended to replace the first), or augmentation (the second augments the first). 

Algorithm~\ref{alg:overaug} describes our approach. Given a chronologically ordered set of ST-slices, we examine each pair of slices $(s_{i},s_{j})$, $j>i$, and determine overwriting and augmentation based on stroke area overlap. This is based on the assumption that two slices that capture an overwrite are likely to have extensive area overlap, as measured by the intersection of the convex hulls of their respective stroke sets. (The intersection is measured in either direction -- $s_{i}$ as percent of $s_{j}$ and vice versa -- if one does not wholly contain the other, and as the smaller as a percentage of the larger if one does contain the other.) Augmentations, by contrast, are likely to have small area overlap (e.g., adding just a ``hat'' or a ``foot'' to a 1). We empirically set the threshold values based on training data ($\theta_{1}$ is 60\% and $\theta_{2}$ is 5\%).

Strokes in a slice deemed overwritten are classified for future reference, then removed from further consideration, because the user replaced them. We classify the strokes using the sketched symbol recognizer of \cite{ouyang2009visual}, trained on a set of properly segmented clock numerals drawn by healthy participants. 

A slice deemed to be an augmentation is merged with its corresponding slice(s). 

The result at this point is a set of slices that we believe represent what the subject intended the final drawing to show, with overwritten ink analyzed and removed.

\subsection{Clock Numeral Classification}
\begin{figure}[tp]
	\begin{center}
		\includegraphics[width=.35\textwidth]{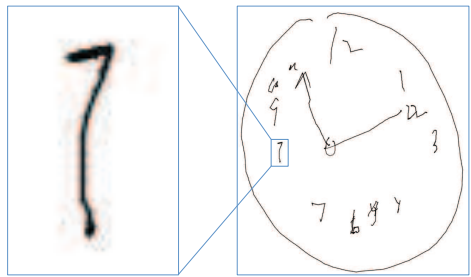}
	\end{center}	
	\caption{Context can be crucial in clock numeral interpretation. When considered in context, the number on the left side is interpreted by trained analysts as an 8, because it sits between the 7 and 9.}
	\label{fig:fig4}
\end{figure}

As has long been recognized, context can be crucial in interpretation. Consider the digit in Fig.~\ref{fig:fig4}. It is clearly ambiguous and could be a 1 or a 7. Yet when considered in context it is interpreted by trained analysts as an 8 (it sits between the 7 and 9). We want our system to have a similar ability to consider more than just the appearance of each set of pen strokes. 

We take context into account by formulating the problem as structured prediction, and use a conditional random field (CRF)~\cite{lafferty2001crf} that provides one way to combine information about what pen strokes look like with contextual information such as where they are in the clock, what's on either side of them angularly, etc. 

We could encode the spatial relationship between the $n$ digit slices using a circular chain structure, mimicking the annular area in a clock face. But this loopy structure makes it difficult to perform inference in the CRF model, as it requires performing approximate inference using methods such as alpha expansion~\cite{szummer2008learning} or mean field approximation~\cite{koltun2011efficient}. 

In response, we take advantage of the fact that our graph is a simple circular chain and break it right after the north slice of the clock face (often the numeral 12), creating a loop-free linear chain graph. We then append to each slice the adjacent slices on either side of it, so that $x_{i} = [x_{i-1};\: x_{i};\: x_{i+1}]$. Each slice feature vector now has a dimension three times the original size. This combination of breaking the circle and concatenating feature vectors permits the use of an efficient belief propagation algorithm~\cite{pearl1982reverend}, while still taking account of spatial context. 

We define the CRF with a singleton term that learns the \textit{appearance} of digits as the compatibility between a digit label $y_{i}$ and the feature vector $x_{i}$ for the $i$-th slice, and a pairwise term that captures the \textit{context} as the compatibility between two angularly consecutive digit labels $y_{i-1}$ and $y_{i}$. For each ST-slice the CRF returns a 12-element vector indicating the probability of each numeral label for that slice.

Training of the CRF follows the standard gradient descent approach ~\cite{lafferty2001crf}. To avoid over-fitting, we used an $l_{2}$ regularization with its scale term set at 0.01, chosen based on 10-fold cross validation. 

To determine the appropriate feature representations to use, we trained and tested 12 CRF classifiers with different combinations of features, including the 24 x 24 feature images from the \cite{ouyang2009visual} symbol recognizer, the angular position of the slice, and the number of pen strokes in the slice. We review the results in Section~\ref{sec:exp}.

\subsection{Repairs}

\begin{algorithm}[tp]
\caption{Repair Strategy for Under-segmentation}
\label{alg:underseg}
\begin{algorithmic}[1]
\State \algorithmicrequire{Set of ST-slices $\mathcal{S}$, ST-slice $s$}
\Procedure{Repair\textendash Undersegmentation}{$\mathcal{S}, s$}
  \State // Find the best scoring partition of $s$
  \State $\mathcal{P} \leftarrow$ partitions of $s$ into sets of chronological strokes
  \For{each partition $P_i \in \mathcal{P}$}
    \For{each set of strokes $p_j \in P_i$}
      \State score of $p_j$ $\leftarrow$ recognize($p_j$)
    \EndFor
  \EndFor
  \State $P^* \leftarrow$ partition with the highest average score
  \State score of $s$ $\leftarrow$ recognize($s$)
  \If{the highest average score $>$ score of $s$}
  \State // Find non-overlapping subset of slices
  \For{$(p_i,p_j) \in P^*, j > i$}
    \If{overlap$(p_i,p_j)$ $> \theta_1$}
      \State $P^* \leftarrow P^* \setminus p_i$
    \EndIf
  \EndFor
  \State // Replace $s$ with $P^*$, preserving stroke order in $\mathcal{S}$
  \State $\mathcal{S} \leftarrow \left( \mathcal{S} \setminus \{s\} \right) \cup P^*$
  \EndIf
\EndProcedure
\end{algorithmic}
\end{algorithm}

\begin{algorithm}[tp]
\caption{Repair Strategy for Over-segmentation}
\label{alg:overseg}
\begin{algorithmic}[1]
\State \algorithmicrequire{Set of ST-slices $\mathcal{S}$, consecutive ST-slices $s_i, s_j$}
\Procedure{Repair\textendash Oversegmentation}{$\mathcal{S},s_i, s_j$}
  \State merged score   $\leftarrow$ recognize(merge$(s_i,s_j)$)
  \State original score $\leftarrow$ avg(recognize($s_i$),recognize($s_j$))
  \If{merged score $>$ original score}
	\State $s_i \leftarrow \mbox{merge}(s_i,s_j)$
    \State $\mathcal{S} \leftarrow \mathcal{S} \setminus \{s_{j}\}$
  \EndIf
\EndProcedure
\end{algorithmic}
\end{algorithm}

\begin{figure}[tp]
	\begin{center}
		\includegraphics[width=.47\textwidth]{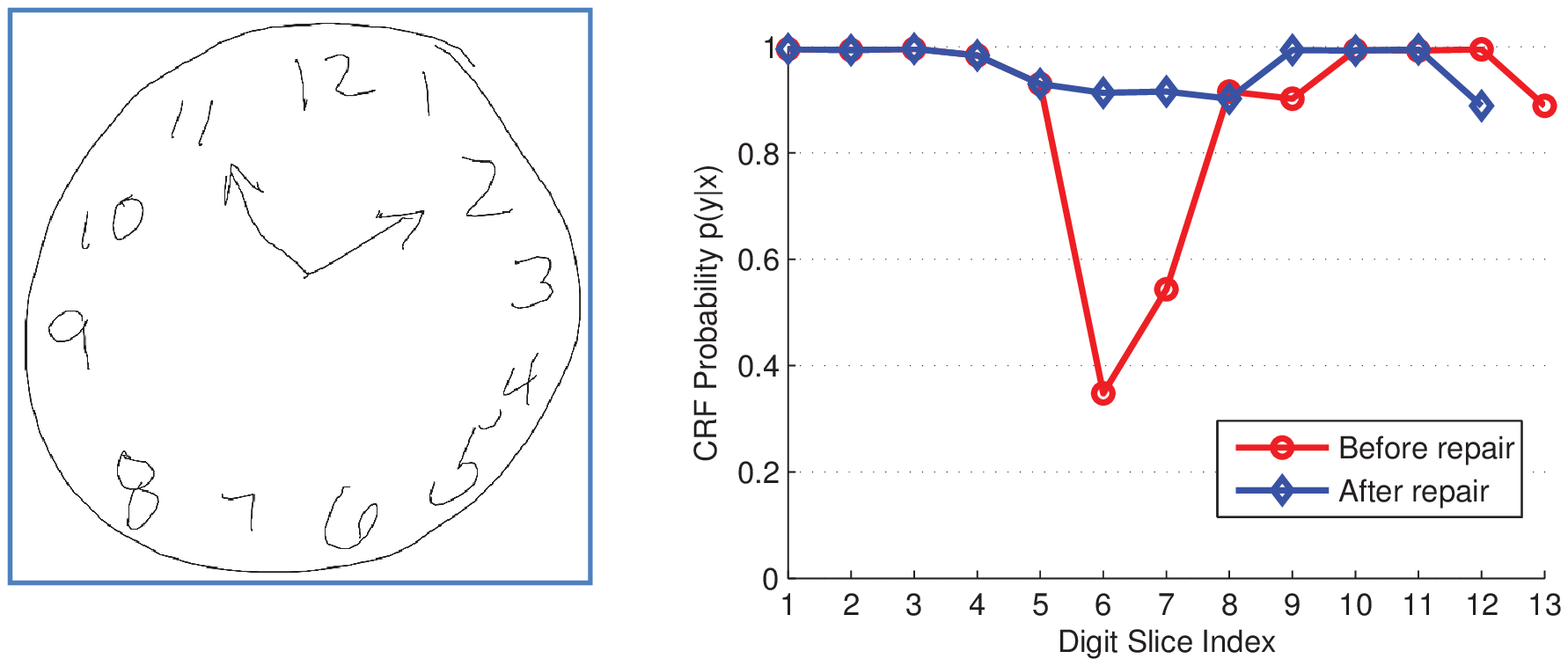}
	\end{center}	
	\caption{A clock drawing sample and the corresponding CRF probabilities before and after applying our repair strategy for over-segmentation. The 5 is initially over-segmented into the 6th and the 7th digit slices, but later corrected.}
	\label{fig:fig5}
\end{figure}

The ST-slice mechanism is of course not perfect: it will under-segment in some cases and over-segment in others. In Fig.~\ref{fig:fig1}(b), the 2 overwritten with the 4 (at the 4 position) produces an under-segmentation, because the 2 was immediately overwritten with the 4, yielding a single ST-slice with all three strokes in it. Conversely, ST-slices can over-segment in circumstances like those in Fig.~\ref{fig:fig5}, described below.

In response, the next step is to examine the posterior probabilities provided by the CRF for the highest ranking interpretation for each slice in sequence, and detect ``valleys'' in this score as a way of finding places where the stroke interpretation may be incorrect (e.g., the 6th and the 7th digit slices) Fig.~\ref{fig:fig5}). We consider a slice to be a valley if its probability score is at least 30\% lower than the score for either of its adjacent slices. Looking at score differences from slice to slice rather than absolute scores adjusts to some extent for clocks with generally badly drawn digits. 

When multiple potential repair sites are found, we start with the repair site whose adjacent slice has the highest score, trying to work from what we are relatively more sure of (the high-scored slice) to help guide the repair (referred to as the ``islands of certainty'' strategy in~\cite{erman1980hearsay}).

After each repair, the entire set of (revised) ST-slices is sent back to the CRF to determine whether the repair improved the interpretation. An improvement in one slice can of course propagate, affecting scores in adjacent slices, leading to an overall improvement and/or the identification of additional repair site candidates. This process continues until no new repair sites are identified.

We developed repair strategies to handle both over-segmentation (e.g., where a single numeral was split into two slices) and under-segmentation (e.g., multiple digits so close angularly they end up in a single ST-slice). Slice properties determine which strategy to apply: oversegmentation is applied to valleys formed by two consecutive slices, while undersegmentation is applied to valleys containing only a single slice that is unusually wide angularly and that has an unusually large number of strokes in it (``unusual'' is defined as more than one standard deviation larger than the average of that property).

Algorithm~\ref{alg:underseg} attempts to deal with under-segmentation by partitioning strokes in a slice into multiple subsets of chronological strokes. For example, given a slice with three strokes $\{1,2,3\}$, we create three partitions $\mathcal{P}=\{P_1,P_2,P_3\}$ where $P_1=\{\{1\},\{2,3\}\}$, $P_2=\{\{1,2\},\{3\}\}$, and $P_3=\{\{1\},\{2\},\{3\}\}$. It then applies the sketched symbol recognizer mentioned in Section~\ref{sec:overaug} \cite{ouyang2009visual}, which has been trained to recognize the twelve clock numerals in isolation, and uses the results to determine whether any of the partitions improves the interpretation. Once the slice is re-segmented using the optimal partition, we remove any subset of strokes if it has been overwritten by strokes in a subsequent subset in that slice (using the same $\theta_1$ = 60\% criterion as in Algorithm~\ref{alg:overaug}). As one example, this deals successfully with the 2 overwritten by a 4 in Fig.~\ref{fig:fig1}(b).

Fig.~\ref{fig:fig5} illustrates an over-segmentation: the top stroke of the 5 is sufficiently far angularly from the midpoint of the base stroke that the two are put in separate slices. Algorithm~\ref{alg:overseg} looks for a valley two slices wide, tries merging them, and evaluates the interpretation of the new set of slices. The graph in Fig.~\ref{fig:fig5} shows the successful repair, with the CRF probabilities demonstrating significant improvement from initial segmentation (red, 13 slices) to the repaired segmentation (blue, 12 slices).

\section{Experiments}
\label{sec:exp}
We used a set of 2,024 clock drawings collected from clinical facilities: 1,654 clocks drawn by healthy participants and 370 clocks randomly selected from mildly cognitively impaired participants. Ground truth labeling was provided by trained analysts, including segmentation and identification of each clock numeral slice.

\subsection{Clock Numeral Identification}
Our final digit identification accuracy ranged from 99.14\% for healthy-user clocks to 92.32\% for impaired-user clocks. For healthy users virtually all errors were in segmentation; using ground truth segmentation labels, digit identification was 99.32\% (see overall performance in Table~\ref{tab:results}). The segmentation errors were typically over-segmentations (e.g., splitting the two digits in a ``12''). Our repair strategies dealt successfully with about half the segmentation errors that occurred, producing the final result of 99.14\%.

Unsurprisingly, impaired clocks had more errors in identification (3.66\%) and segmentation (4.08\%), yielding an overall accuracy of 92.32\%. Errors in these cases typically resulted when a sequence of digits were drawn badly (e.g., the 1,2,3,4,5,6 sequence in Fig.~\ref{fig:fig1} (c)), preventing context from offering sufficient guidance.

Overall our performance compares favorably to the state of the art isolated digit recognition -- an accuracy of 99.79\% on the MNIST dataset \cite{wan2013regularization} -- considering the added difficulties we face (non-isolated digits, 12 labels to assign, some extremely challenging data). It also illustrates the power of context to aid interpretation, something obviously not used in isolated digit recognizers. 

\begin{figure}[t]
	\begin{center}
		\includegraphics[width=.45\textwidth]{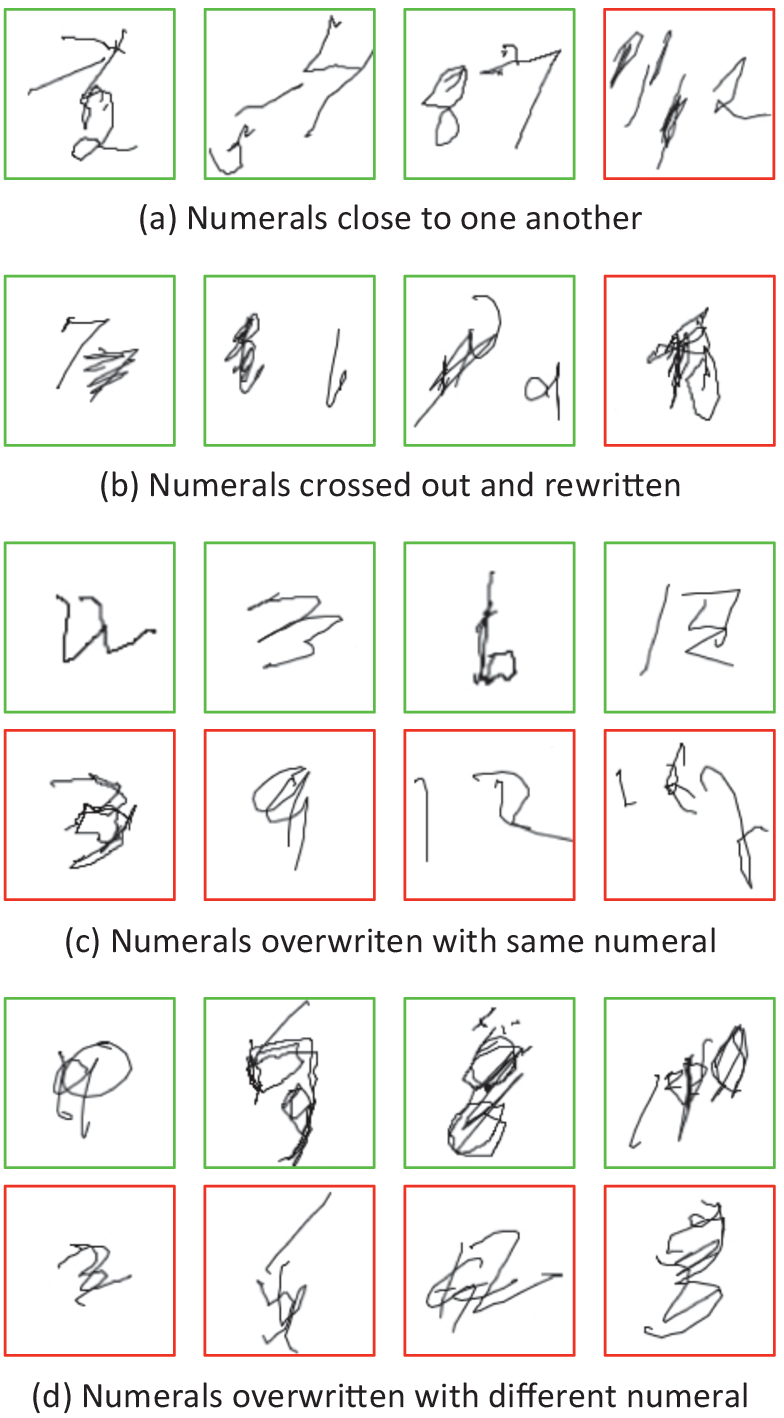}
	\end{center}	
	\caption{Success (green) and failure (red) cases of ST-slices, 
    categorized into four canonical types of complications.}
	\label{fig:st-slice-qualitative}
\end{figure}

{
\begin{table*}[tp]
	\tabulinesep=1.3mm
	\centering
	\begin{tabu}{l|c|c|c|c|c}
    
    \bf{Trained on} & \bf{Input} & \bf{Ang, \# Stk} & 
    \bf{Overall Perf (sd)} & \bf{Perf Healthy (sd)} & \bf{Perf Impaired (sd)} \\ 
    \hline
    Healthy & single  &   & 89.93\% (0.0101) & 93.83\% (0.0103) & 71.03\% (0.0309) \\ 
    Healthy & single  & Y & 87.51\% (0.0068) & 90.51\% (0.0074) & 73.01\% (0.0321) \\ 
    Healthy & concat. &   & 98.43\% (0.0049) & 96.76\% (0.0018) & 92.01\% (0.0243) \\ 
    Healthy & concat. & Y & 99.01\% (0.0045) & 99.90\% (0.0011) & 94.66\% (0.0232) \\ 
    \hline    
    Impaired & single  &   & 75.11\% (0.2420) & 78.43\% (0.0268) & 59.04\% (0.0232) \\ 
    Impaired & single  & Y & 76.76\% (0.0285) & 79.05\% (0.0328) & 65.67\% (0.0234) \\ 
    Impaired & concat. &   & 98.84\% (0.0055) & 99.75\% (0.0024) & 94.45\% (0.0234) \\ 
    Impaired & concat. & Y & 99.05\% (0.0092) & 99.82\% (0.0025) & 95.31\% (0.0429) \\ 
    \hline
    Both & single  &   & 88.96\% (0.0080) & 92.88\% (0.0094) & 69.95\% (0.0224) \\ 
    Both & single  & Y & 88.29\% (0.0137) & 92.18\% (0.0140) & 75.31\% (0.0264) \\ 
    Both & concat. &   & 98.85\% (0.0048) & 99.82\% (0.0018) & 94.13\% (0.0228) \\ 
    Both & concat. & Y & \bf{99.32\%} (0.0028) & \bf{99.93\%} (0.0009) & \bf{96.34\%} (0.0160) \\     
	\end{tabu}
	\caption{We report mean accuracy and standard deviation across 10 folds. We measured the effect of several factors, including choice of training set (healthy, impaired, both), whether feature vectors for the CRF were concatenated for three adjacent slices, and whether the feature vector contained only 24 x 24 feature images, or included slice angular position and stroke count. 
}
	\label{tab:results}
\end{table*}
}

\subsection{Quantifying The Impact of ST-Slices}

Our segmentation method (ST-slices) can deal with overwriting that is otherwise incomprehensible even to humans. To see its effectiveness, we selected 70 clocks in our dataset that contain at least one incomprehensible part resulting from overwriting.

Our ST-slice mechanism separated the layers of ink accurately in 79\% of these cases.  As a comparison, a purely temporal segmenter -- a boosted logistic regression classifier~\cite{friedman2000additive} trained on just the time difference between successive strokes -- performed significantly worse at an accuracy of 66\%. 

Fig.~\ref{fig:st-slice-qualitative} provides some insight into when our technique works well and when it fails. ST-slices performed well on cases where two slices are spatially close (which confused a purely spatial segmenter), but still temporally distinguishable. This handles cases in which two numerals are written near each other but at different times (Fig.~\ref{fig:st-slice-qualitative}a) and cases in which a numeral is crossed out and another written next to it (b). Our technique correctly segmented most of the slices produced in these cases (the green boxes in Fig.~\ref{fig:st-slice-qualitative}a and Fig.~\ref{fig:st-slice-qualitative}b).

Given the complexity of the ink that results from overwriting, none of these would have been handled by existing digit recognition or sketch interpretation techniques.

One source of failure came from immediate overwriting, with either the same (Fig.~\ref{fig:st-slice-qualitative}c) or a different digit (Fig.~\ref{fig:st-slice-qualitative}d), as these strokes end up in the same ST-slice. As noted, we handle this by examining sequential subsets of strokes in the slice, looking for subsets recognizable as a digit. Fig.~\ref{fig:st-slice-qualitative}c and d shows examples of both successes and failures.

A relatively rare problem arises from numerals assembled with non-chronological strokes; our system always fails to correctly (re)group temporally interspersed strokes. These scenarios require determining intent, and are a focus of continued work.

\subsection{Quantifying The Impact of Context}

Given the important role of context in interpretation, particularly in clocks by impaired users, we did a set of experiments to provide a quantitative calibration of the contribution of different context features. 

We trained and tested our digit classifier with 12 different combinations of context features, using as the base feature the 24 x 24 feature images from the \cite{ouyang2009visual} symbol recognizer. To focus on calibrating the contribution of context features, training and testing was done on correctly segmented pen strokes (i.e., isolated clock numerals), using 10-fold cross validation.

We report mean accuracy and standard deviation across the folds. As Table~\ref{tab:results} suggests, adding angle and stroke count alone increases accuracy only a small amount, from 98.85\% to 99.32\% (the last two rows). Concatenating data from adjacent slices provides a more significant performance boost, from 88.96\% to 98.85\% (the second and the fourth to the last row). Not surprisingly, concatenation provided a particularly significant benefit for impaired clocks, where performance improved from 69.95\% to 94.13\%.

In terms of training sets, training on only the impaired clocks typically gave the worst results, even for performance on impaired clocks, perhaps because some digits are so distorted that they are blurring the ability to distinguish more clearly drawn digits.

Overall these results give quantitative evidence for the significant contribution made by considering each digit in the context of the two on either side of it. While not unexpected, it is useful to see that it provides the kind of capability evidenced by human performance, e.g., for the digit in Fig.~\ref{fig:fig4}, which is now properly classified as an 8.

\section{Related Work}
Work in~\cite{kim2013clockme} is also focused on the clock drawing test, but emphasizes interface design, seeking to create a tablet-based system that is usable by both subjects and clinicians. It does basic digit recognition but does not report dealing with distorted or over-written digits, or the other complexities described here.

Digit recognition has a long history, for both online and offline capture: \cite{plamondon2000online} provides a survey, recent work on convolutional neural networks offers current performance levels~\cite{ciresan2012multi,wan2013regularization}. While we need a good digit recognizer, our larger focus is on the identification of digits in context in challenging conditions (e.g., distorted by impairment, crossed out, overwriting, etc.), which has not been studied extensively.

Our work on ST-slices builds on and extends a long history of using temporal information in sketch understanding, where it has been used in a variety of ways, including stroke segmentation (e.g. \cite{ouyang2009learning}) and interpretation \cite{sezgin2008sketch,taele2008using}. While previous work has explored over-tracing (writing the same thing on top of itself, often for emphasis, e.g., \cite{sezgin2004handling}), overwriting (writing to replace) appears to be rarely tackled. \cite{hurst1998error} take on the task, but set it in the context of interactive handwriting recognition, in which the user and the system work cooperatively, which is inappropriate for our task. Work in \cite{dickmann2010sketch} combines spatial and temporal information to do stroke segmentation, but appears to deal only with consecutive strokes.

Spatio-temporal information have also been used together in a variety of circumstances, as for example dense trajectories~\cite{wang2013dense} and C3D~\cite{tran2014c3d} for video analysis in the computer vision community. Our visual feature representation from ~\cite{ouyang2009visual} can be thought of as 2D convolutional features obtained from pre-defined filters; using 3D convolutional features~\cite{tran2014c3d} to represent our pen stroke data could give us extra performance boost. This is the focus of our ongoing work.

\section{Conclusions}
Our sketch interpretation system balances visual appearance and context, enabling it to handle some of the complex phenomena found in drawings produced during a real-world task performed by a wide range of naive users behaving as they do ordinarily. 

We demonstrate our system on detecting and classifying clock numerals from drawings created by subjects taking the digital Clock Drawing Test. We report performance on clock numeral detection and classification at the 96\%--99\% level even for sketches drawn by those with impairments, calibrating and demonstrating the utility of different forms of context. We show that a novel representation -- ST-slices -- and the processing that accompanies it enables unpeeling and interpreting the layers of ink that result when people overwrite, giving our system a novel ability to classify symbols in a drawing even though they have been over-written or crossed out.

As the performance figures suggest, the system works well on clock numeral isolation and detection for clocks from both healthy and impaired users. Even so there are several avenues of improvement to be explored. 

Perhaps the most important is to explore alternatives to the sequential character of our system (segmentation and recognition). Human segmentation of strokes is clearly guided in part by understanding the drawing, i.e., segmentation and interpretation work simultaneously. Recent research has shown that convolutional neural networks combined with region proposals are very effective at simultaneous detection and recognition of visual objects from images~\cite{girshick2014rich}. Applying techniques like this to our domain would require additional steps to deal with cross-outs and over-writing. We are looking into ways to accomplish this and expect it to reduce the role of segmentation as a major source of clock numeral identification error.

We have focused here on clock numeral identification because it is the most difficult part of the task, but full sketch interpretation for the clock means recognizing all the other elements (hands, tick marks, etc.). We have a start on this but there is considerably more to do.

Where possible we compared our work to the state of the art, e.g., its performance on healthy clocks, which is comparable to the best isolated digit recognizers on the MNIST dataset~\cite{wan2013regularization}. But an important, novel part of our system is space-time slicing for sketch recognition, and its ability to handle the cross-outs and over-writing that occur in ordinary drawing and writing. Since there is yet no standard dataset or alternative program to compare directly against for this part of the system, we provided the qualitative analysis (Fig.~\ref{fig:st-slice-qualitative}) that explains in detail when space-time slicing works well and when it fails. 

While ST-slices have been applied here to a specific task, the concept is more generally applicable to any sketch understanding task where layers of ink are encountered and need to be unraveled. More generally, they are a combination of the spatial and temporal properties of pen strokes and as such are applicable to a variety of sketch recognition domains, e.g., drawings of chemical structures~\cite{ouyang2011chemink}.

The general idea of using context to guide signal understanding is of course both known and widely applicable. This work provides a calibration of its power on a specific task and more generally shows how spatial and temporal information can be captured in a way that is powerful on this specific task, yet general enough to be applicable to a variety of other hand drawing interpretation tasks.

\section*{Acknowledgment}
This work was supported in part by NSF Grant IIS-1404494 and by the REW Research and Education Institution.

\bibliographystyle{named}
\bibliography{ijcai16}
\end{document}